\documentclass[letterpaper, 10 pt, conference]{ieeeconf}  

\IEEEoverridecommandlockouts                              

\usepackage{amsmath} 
\usepackage{amsfonts}
\usepackage{booktabs}
\usepackage{multirow}
\usepackage{graphicx}
\usepackage{xcolor}

\title{\LARGE \bf
ART: Adaptive Relational Transformer for Pedestrian Trajectory Prediction with Temporal-Aware Relations
}

\author{Ruochen Li\textsuperscript{\rm 1}, Ziyi Chang\textsuperscript{\rm 1}, Junyan Hu\textsuperscript{\rm 1}, Jiannan Li\textsuperscript{\rm 2}, Amir Atapour-Abarghouei\textsuperscript{\rm 1}, Hubert P. H. Shum\textsuperscript{\rm 1*}\thanks{*Corresponding author}
\thanks{$^{1}$Ruochen Li, Ziyi Chang, Junyan Hu, Amir Atapour-Abarghouei and Hubert P. H. Shum are with Department of Computer Science, Durham University, UK {\tt\small \{ruochen.li, ziyi.chang, junyan.hu, amir.atapour-abarghouei, hubert.shum\}@durham.ac.uk}}%
\thanks{$^{2}$Jiannan Li is with the School of Computing and Information Systems, Singapore Management University, Singapore
        {\tt\small  jiannanli@smu.edu.sg}}%
}

\begin{document}

\maketitle
\thispagestyle{empty}
\pagestyle{empty}

\begin{abstract}

Accurate prediction of real-world pedestrian trajectories is crucial for a wide range of robot-related applications. Recent approaches typically adopt graph-based or transformer-based frameworks to model interactions. Despite their effectiveness, these methods either introduce unnecessary computational overhead or struggle to represent the diverse and time-varying characteristics of human interactions. In this work, we present an Adaptive Relational Transformer (ART), which introduces a Temporal-Aware Relation Graph (TARG) to explicitly capture the evolution of pairwise interactions and an Adaptive Interaction Pruning (AIP) mechanism to reduce redundant computations efficiently. Extensive evaluations on ETH/UCY and NBA benchmarks show that ART delivers state-of-the-art accuracy with high computational efficiency.
\end{abstract}

\section{INTRODUCTION}

Pedestrian trajectory prediction aims to forecast future locations of pedestrians given observed trajectories. It is a key component of human–robot interaction \cite{akabane2021pedestrian2,huang2022pedestrian3}, autonomous driving systems \cite{bai2015pomdpintro,luo2018porca}, and smart surveillance infrastructure \cite{li2025unifiedTraj,li2025bpsgcn}. Accurate prediction of real-world pedestrian trajectories is crucial for robot-related applications, including human-aware robot navigation \cite{bhaskara2024trajectory1,chen2018robotnavigation1} and collision avoidance systems \cite{yang2024ia,li2025vite}. However, pedestrian trajectory prediction remains challenging due to the inherent stochasticity of motion behaviors and social interactions.

Early approaches to pedestrian trajectory prediction modeled social interactions using pooling-based strategies that aggregate information from nearby pedestrians through fixed spatial windows \cite{Alexandre2016lstm,gupta2018socialgan} or occupancy maps \cite{xue2018ss-lstm,guo2022end2end}. To enable explicit and expressive interaction modeling, two main directions have emerged. On the one hand, graph-based methods encode pedestrians as nodes and model their pairwise interactions as edges, enabling structured message-passing via graph neural networks \cite{ruochen2022multiclassSGCN,li2025unifiedTraj,xu2023eqmotion,li2026vite}. Specifically, these methods propagate interaction features via a message-passing paradigm over graph structures to model social influence among pedestrians. On the other hand, transformer-based approaches \cite{lee2024mart,shi2023tutr,yu2020spatio,yuan2021agentformer} leverage self-attention mechanisms \cite{vaswani2017transformer} to model interactions by dynamically weighting the influence of surrounding agents across both spatial and temporal dimensions, providing a flexible framework for capturing long-range dependencies and complex motion patterns.

Despite these advances, previous methods either incur redundant computation or fail to capture the heterogeneous and evolving nature of pedestrian interactions. Pedestrian trajectories contain rich spatial-temporal information, yet existing methods \cite{lee2024mart,xu2023eqmotion,shi2023tutr,qiao2024category} construct pair-wise relations based on temporally aggregated representations, where trajectory sequences are compressed into static node features and relational reasoning is performed on these representations, thereby overlooking the dynamic evolution of interactions and time-varying strength. In addition, many methods construct social graphs using uniform neighbor selection, such as fully connected graphs \cite{lee2024mart,qiao2025geometric} or top-\textit{k} based neighbor selection \cite{li2025unifiedTraj,xu2023eqmotion}, which ignore the heterogeneous nature of pedestrian interactions.

\begin{figure}[t]
\centering
\includegraphics[width=0.72\columnwidth]{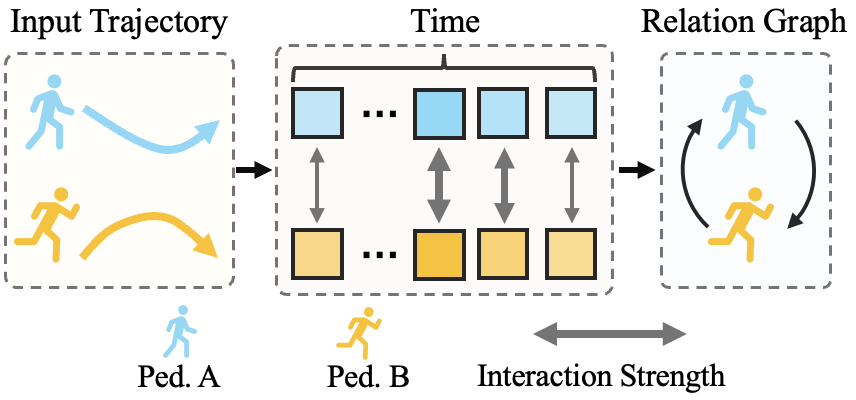} 
\caption{Framework overview. Framework overview. The relation between pedestrians is inferred from the temporal evolution of their pairwise interactions over the observed history.}
\label{fig:teaser_image}
\vspace{-4mm}
\end{figure}

\begin{figure*}[th]
  \centering
  \includegraphics[width=0.90\linewidth]{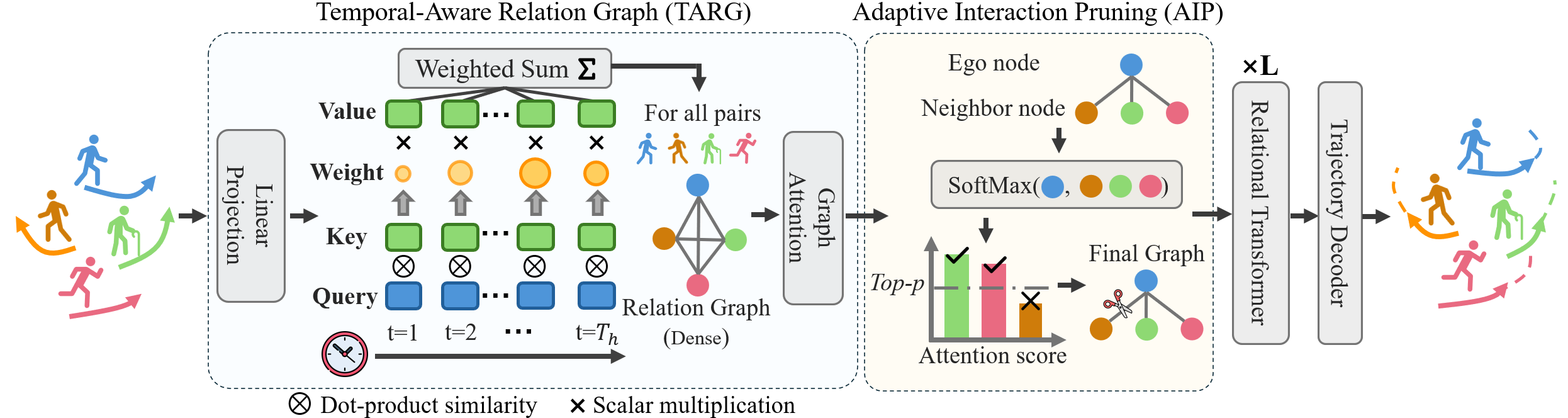}
  \caption{Overview of ART. \textbf{Left:} Temporal-Aware Relation Graph (TARG) leverages pairwise attention to model agent interactions across time steps, assigning higher weights to informative moments. \textbf{Right:} Adaptive Interaction Pruning (AIP) uses top-\textit{p} filtering to adaptively retain informative neighbors based on cumulative interaction strength, producing a sparsified graph for trajectory prediction.}
  \label{fig:overview}
\end{figure*}

In this work, we propose an Adaptive Relational Transformer (ART), a lightweight Transformer-based framework for pedestrian trajectory prediction. First, to explicitly capture the temporal evolution of pairwise interactions, we introduce a Temporal-Aware Relation Graph (TARG). Unlike existing methods that compress multi-step trajectories into static embeddings before computing pairwise 
relations, TARG leverages pairwise attention \cite{ha2022learningrain} to model agent interactions at each time step and aggregates them via learnable weights, providing a temporally-aware relation modeling where informative time steps (e.g., moments of close proximity or directional change) receive higher weights, while less relevant moments contribute minimally to the final pairwise relations (see Figure~\ref{fig:teaser_image}). Second, to reduce redundancy in dense interaction graphs, we propose Adaptive Interaction Pruning (AIP) via top-\textit{p} filtering, which adaptively selects informative neighbors based on interaction strength \cite{lin2025twilight}, rather than enforcing a fixed top-\textit{k} neighborhood. For each agent, only neighbors whose attention weights jointly exceed a threshold \textit{p} are retained, resulting in an adaptively sparsified interaction graph that retains only informative neighbors. 

The proposed ART achieves state-of-the-art (SOTA) results on both the ETH/UCY \cite{pellegrini2009ETH,Lerner2007UCY} and NBA SportVU \cite{mao2023Leapfrog} benchmarks while maintaining superior computational efficiency over prior approaches. The main contributions are:
\begin{itemize}
    \item We propose an Adaptive Relational Transformer (ART), a lightweight Transformer-based framework for pedestrian trajectory prediction that explicitly models social interactions while maintaining computational efficiency.
    \item We introduce a Temporal-Aware Relation Graph (TARG) that explicitly models pairwise pedestrian interactions by accounting for the relative importance of different time steps.
   \item We propose Adaptive Interaction Pruning (AIP) via top-\textit{p} filtering to adaptively sparsify dense interaction graphs based on cumulative interaction strength, reducing redundancy and improving adaptivity.
\end{itemize}

\section{METHOD}

\subsection{Problem Formulation}
Pedestrian trajectory prediction aims to infer pedestrians' future movements from their observed trajectories. Consider a scene containing $M$ pedestrians observed over $T_h$ time steps. The historical trajectory of pedestrian $i$ is represented as $\mathbf{p}_j^{h} = \{ (x^t_{i}, y^t_{i}) \mid (x^t_{i},y^t_{i})\in \mathbb{R}^2, t = 1, \dots, T_h \}$, where $(x^t_{i}, y^t_{i})$ denotes the 2D spatial coordinates at time step $t$. The corresponding ground-truth (GT) future trajectory over a prediction horizon of $T_f$ time steps is denoted as $\mathbf{p}_j^{f} = \{ (x^t_{i}, y^t_{i}) \mid t = T_h + 1, \dots, T_h + T_f \}$. By stacking trajectories of all pedestrians, the observed and GT trajectories can be expressed as $\mathbf{P}^{h} \in \mathbb{R}^{M \times T_h \times 2}$ and $\mathbf{P}^{f} \in \mathbb{R}^{M \times T_f \times 2}$. The training objective minimizes the discrepancy between the predicted trajectories $\hat{\mathbf{P}}^{f}$ and the ground-truth $\mathbf{P}^{f}$.

\subsection{Temporal-Aware Relation Graph Construction}

Existing graph construction methods compress $\mathbf{P}_h \in \mathbb{R}^{M \times T_h \times 2}$ into static embeddings before computing pairwise weights, losing temporal information about when interactions occur. As shown in Figure~\ref{fig:overview} (Left), we propose Temporal-Aware Relation Graph (TARG), which preserves temporal attribution by applying attention over observation 
sequences before spatial aggregation.

We encode the observed trajectories into high-dimensional features. Given $\mathbf{P}_h \in \mathbb{R}^{M \times T_h \times 2}$, we apply a linear projection with positional encoding to obtain temporal node features:
\begin{equation}
    \mathbf{H} = \text{PosEnc}(\mathbf{W}_{\text{in}} \mathbf{P}_h) \in \mathbb{R}^{M \times T_h \times d},
\end{equation}
where $\mathbf{W}_{\text{in}} \in \mathbb{R}^{d \times 2}$ is a learnable projection matrix, $d$ is the hidden dimension, and $\text{PosEnc}(\cdot)$ denotes sinusoidal positional encoding \cite{vaswani2017transformer} that injects temporal information.

To capture time-varying interactions between pedestrian pairs, we reshape $\mathbf{H}$ from $\mathbb{R}^{M \times T_h \times d}$ to $\mathbb{R}^{T_h \times M \times d}$ and apply multi-head attention \cite{vaswani2017transformer} over the temporal dimension. For each attention head 
$h \in \{1, \dots, H\}$, we have:
\begin{equation}
    \mathbf{Q}^h = \mathbf{H} \mathbf{W}_Q^h, \quad 
    \mathbf{K}^h = \mathbf{H} \mathbf{W}_K^h, \quad 
    \mathbf{V}^h = \mathbf{H} \mathbf{W}_V^h,
\end{equation}
where $\mathbf{W}_Q^h, \mathbf{W}_K^h, \mathbf{W}_V^h \in \mathbb{R}^{d \times d_h}$ are learnable weight matrices and $d_h = d / H$ is the dimension per head. For each pair of pedestrians $(i, j)$, we compute time-resolved attention scores by applying attention across all time steps:
\begin{equation}
    \alpha_{ij}^h(t) = \frac{\exp\left(\frac{\mathbf{Q}^h_i(t)^\top \mathbf{K}^h_j(t)}{\sqrt{d_h}}\right)}{\sum_{t'=1}^{T_h} \exp\left(\frac{\mathbf{Q}^h_i(t')^\top \mathbf{K}^h_j(t')}{\sqrt{d_h}}\right)}, \quad t = 1, \dots, T_h,
\end{equation}
where $\alpha_{ij}^h(t) \in [0, 1]$ represents the attention weight for pedestrian pair $(i, j)$ at time step $t$ under head $h$, and $\sum_{t=1}^{T_h} \alpha_{ij}^h(t) = 1$. These temporally-aware scores explicitly capture when each interaction becomes significant, preserving temporal attribution throughout the graph construction.

The attention scores are used to aggregate value representations across time:
\begin{equation}
    \mathbf{R}_{ij}^h = \sum_{t=1}^{T_h} \alpha_{ij}^h(t) \cdot \mathbf{V}^h_j(t) \in \mathbb{R}^{d_h},
\end{equation}
where $\mathbf{R}_{ij}^h$ is the aggregated relation representation for pedestrian pair $(i, j)$ under head $h$. By concatenating outputs from all heads and applying a linear projection, we obtain the pairwise relation features $\mathbf{R}_{ij}$.

Finally, we compute edge weights from the pairwise relation features $\mathbf{R}_{ij}$. Following the graph attention mechanism \cite{petar2018GAT}, we apply a learnable function to produce edge weights:
\begin{equation}
    w_{ij} = \sigma\left( \mathbf{a}^\top \left[ \mathbf{R}_{ij} \| \mathbf{R}_{ji} \right] + b \right),
\end{equation}
where $\mathbf{a} \in \mathbb{R}^{2d}$ and $b \in \mathbb{R}$ are learnable parameters, $\sigma(\cdot)$ is the sigmoid activation function, and the concatenation of $\mathbf{R}_{ij}$ and $\mathbf{R}_{ji}$ captures bidirectional interactions. The adjacency matrix $\mathbf{W}\in\mathbb{R}^{M\times M}$ with entries $w_{ij}$ defines the temporal-aware relation graph.

\subsection{Adaptive Interaction Pruning}

As discussed in the previous section, TARG produces an adjacency matrix $\mathbf{W}$ capturing pairwise interactions. However, modeling all $M(M-1)$ relations incurs $\mathcal{O}(M^2)$ computational complexity and introduces noise from spurious interactions. To reduce computational redundancy while enabling adaptive neighbor selection, we propose Adaptive Interaction Pruning (AIP) (Figure~\ref{fig:overview} (Right)) via top-$p$ filtering \cite{lin2025twilight}, which allows each pedestrian to discover a personalized neighbor set based on attention distributions. For each target pedestrian $i$, we rank potential neighbors $\{j \mid j \neq i\}$ by their edge weights in descending order. Let $\pi_i$ denote the permutation that sorts edge weights:
\begin{equation}
    w_{i, \pi_i(1)} \geq w_{i, \pi_i(2)} \geq \cdots \geq w_{i, \pi_i(M-1)}.
\end{equation}
We then compute the cumulative sum of sorted weights:
\begin{equation}
    C_i(k) = \sum_{j=1}^{k} w_{i, \pi_i(j)}, \quad k = 1, \dots, M-1.
\end{equation}
Given threshold $p \in (0, 1]$, we retain the minimal set of neighbors whose cumulative weight exceeds $p$:
\begin{equation}
    k_i^* = \min \left\{ k \mid \frac{C_i(k)}{C_i(M-1)} \geq p \right\},
\end{equation}
where $k_i^*$ is the effective neighborhood size for pedestrian $i$. The pruned adjacency matrix $\tilde{\mathbf{W}} \in \mathbb{R}^{M \times M}$ is obtained by:
\begin{equation}
    \tilde{w}_{ij} = \begin{cases}
        w_{ij} & \text{if } j \in \{\pi_i(1), \dots, \pi_i(k_i^*)\}, \\
        0 & \text{otherwise}.
    \end{cases}
\end{equation}
Finally, we renormalize the weights for each pedestrian as:
\begin{equation}
\tilde{w}_{ij} = \frac{\tilde{w}_{ij}}{\sum_{j'=1}^{M} \tilde{w}_{ij'}}.
\end{equation}

This design adaptively determines neighborhood size based on learned interaction strength, resulting in a sparsified yet informative interaction graph that improves both efficiency and robustness.

\subsection{Relational Transformer and Trajectory Decoding}
Given the pruned adjacency matrix $\tilde{\mathbf{W}} \in \mathbb{R}^{M \times M}$ from AIP, we employ $L$ layers of Relational Transformer (RT) \cite{lee2024mart,diao2023relational} to refine node representations through edge-aware attention. Let $\mathbf{h}_i^{(0)} \in \mathbb{R}^d$ denote the initial node feature for pedestrian $i$ obtained from TARG. 

At each layer $l \in \{1, \dots, L\}$, RT updates both node and edge features. For node updates, it applies multi-head attention modulated by edge features 
$\mathbf{e}_{ij}^{(l-1)}$:
\begin{equation}
    \alpha_{ij}^{(l)} = \text{softmax}_j \left( \frac{(\mathbf{Q}_i^{(l)})^\top (\mathbf{K}_j^{(l)} + \mathbf{K}_{\mathbf{e}_{ij}}^{(l)})}{\sqrt{d_h}} \right),
\end{equation}
where $\mathbf{Q}_i^{(l)}, \mathbf{K}_j^{(l)}$ are node-based query and key projections, $\mathbf{K}_{\mathbf{e}_{ij}}^{(l)}$ is the edge-based key, and the attention is computed only over pruned neighbors (where $\tilde{w}_{ij} > 0$). Node features are then updated via:
\begin{equation}
    \mathbf{h}_i^{(l)} = \mathbf{h}_i^{(l-1)} + \text{FFN}\left( \sum_{j: \tilde{w}_{ij} > 0} \alpha_{ij}^{(l)} (\mathbf{V}_j^{(l)} + \mathbf{V}_{\mathbf{e}_{ij}}^{(l)}) \right),
\end{equation}
where $\mathbf{V}_j^{(l)}$ and $\mathbf{V}_{\mathbf{e}_{ij}}^{(l)}$ are value projections, and FFN denotes a feed-forward network with residual connection. Edge features are updated by aggregating source and target node information:
\begin{equation}
    \mathbf{e}_{ij}^{(l)} = \mathbf{e}_{ij}^{(l-1)} + \text{MLP}\left([\mathbf{h}_i^{(l)} \| \mathbf{h}_j^{(l)} \| \mathbf{e}_{ij}^{(l-1)} \| \mathbf{e}_{ji}^{(l-1)}]\right),
\end{equation}
where $\|$ denotes concatenation and the bidirectional edge features capture symmetric interactions. This edge-aware mechanism allows RT to leverage the pruned graph structure from AIP while refining pairwise relations. We use $L=1$ layer in our experiments for efficiency.

For trajectory prediction, we concatenate the initial node features $\mathbf{h}_i^{(0)}$ and the refined features $\mathbf{h}_i^{(L)}$ after RT encoding, then feed them into $K$ parallel prediction heads to obtain trajectories following \cite{xu2023eqmotion,lee2024mart}:
\begin{equation}
    \hat{\mathbf{p}}^{f}_{i,k} = \text{MLP}_k([\mathbf{h}_i^{(0)} \| \mathbf{h}_i^{(L)}]) \in \mathbb{R}^{T_f \times 2}, \quad k = 1, \dots, K,
\end{equation}
where $\|$ denotes concatenation and $\hat{\mathbf{p}}^{f}_{i,k}$ represents the $k$-th predicted future trajectory for pedestrian $i$. 

Given $K$ predicted trajectories $\{\hat{\mathbf{p}}_{f}^{i,k}\}_{k=1}^K$ for each pedestrian $i$, we adopt the best-of-$K$ training strategy by selecting the trajectory with minimum $\ell_2$ distance to the ground truth:
\begin{equation}
    \mathcal{L} = \frac{1}{M T_f} \sum_{i=1}^M \sum_{t=1}^{T_f} \min_{k \in \{1, \dots, K\}} \left\| \mathbf{p}^{f,t}_{i} - \hat{\mathbf{p}}^{f,t}_{i,k} \right\|_2,
\end{equation}
where $\mathbf{p}^{f,t}_{i} = (x^t_i, y^t_i)$ is the ground-truth position of pedestrian $i$ at time $t$, and $\hat{\mathbf{p}}^{f,t}_{i,k}$ is prediction from the $k$-th head.

\begin{table*}[t]
\centering
\caption{Quantitative result on the ETH/UCY dataset. Metrics are $\min\mathrm{ADE}_{20}$/$\min\mathrm{FDE}_{20}$. Best and second-best results are marked in \textbf{bold} and \underline{underline}.}
\begingroup
\setlength{\tabcolsep}{1mm}
\small
\begin{tabular}{c|ccccccccc|c}
\toprule
\multicolumn{11}{c}{\textbf{ETH/UCY Dataset}} \\
\midrule
Subset 
& GroupNet & MemoNet & MID & NPSN & EqMotion & ET & LED & SingularTraj & MART & Ours \\
& \cite{Xu2022GroupNetMH} & \cite{xu2022remembermemo} & \cite{gu2022stochastic} 
& \cite{bae2022NPSN} & \cite{xu2023eqmotion} & \cite{bae2023eigentrajectory} 
& \cite{mao2023Leapfrog} & \cite{bae2024singulartrajectory} 
& \cite{lee2024mart} &  \\
\midrule
ETH    
& 0.46/0.73 & 0.40/0.61 & 0.39/0.66 & \underline{0.36}/0.59 & 0.40/0.61 & \underline{0.36}/0.53 & 0.39/0.58 & \textbf{0.35}/\textbf{0.42} & \textbf{0.35}/\underline{0.47} & \textbf{0.35}/\underline{0.47} \\

HOTEL  
& 0.15/0.25 & \textbf{0.11}/\textbf{0.17} & 0.13/0.22 & 0.16/0.25 & \underline{0.12}/\underline{0.18} & \underline{0.12}/0.19 & \textbf{0.11}/\textbf{0.17} & 0.13/0.19 & 0.14/0.22 & 0.13/0.21 \\

UNIV   
& 0.26/0.49 & 0.24/0.43 & \underline{0.22}/0.45 & 0.23/\underline{0.39} & 0.23/0.43 & 0.24/0.43 & 0.26/0.43 & 0.25/0.44 & 0.25/0.45 & 0.24/0.43 \\

ZARA1  
& 0.21/0.39 & \underline{0.18}/0.32 & \textbf{0.17}/0.30 & \underline{0.18}/0.32 & \underline{0.18}/0.32 & 0.19/0.33 & \underline{0.18}/\textbf{0.26} & 0.19/0.32 & \textbf{0.17}/0.29 & \textbf{0.17}/\underline{0.28} \\

ZARA2  
& 0.17/0.33 & 0.14/0.24 & \underline{0.13}/0.27 & 0.14/0.25 & \underline{0.13}/0.23 & 0.14/0.24 & \underline{0.13}/\underline{0.22} & 0.15/0.25 & \underline{0.13}/\underline{0.22} & \textbf{0.12}/\textbf{0.21} \\
\midrule
AVG  
& 0.25/0.44 & \underline{0.21}/0.35 & \underline{0.21}/0.38 & \underline{0.21}/0.36 & \underline{0.21}/0.35 & \underline{0.21}/0.34 & \underline{0.21}/\underline{0.33} & \underline{0.21}/\textbf{0.32} & \underline{0.21}/\underline{0.33} & \textbf{0.20}/\textbf{0.32} \\
\bottomrule
\end{tabular}
\label{eth_results}
\endgroup
\end{table*}

\begin{table*}[t]
\centering
\caption{Quantitative result on the NBA dataset. Metrics are $\min\mathrm{ADE}_{20}$/$\min\mathrm{FDE}_{20}$. Best and second-best results are marked in \textbf{bold} and \underline{underline}.}
\begingroup
\setlength{\tabcolsep}{1.5mm}
\small
\begin{tabular}{l|ccccccccc|c}
\toprule

\multicolumn{11}{c}{\textbf{NBA Dataset}} \\
\midrule
Time
& STAR & GroupNet & MemoNet & MID & NPSN & DynGroupNet & LED & SingularTraj & MART & Ours \\
& \cite{yu2020spatio} & \cite{Xu2022GroupNetMH} & \cite{xu2022remembermemo}
& \cite{gu2022stochastic} & \cite{bae2022NPSN}
& \cite{xu2024dynamicGroupNet} & \cite{mao2023Leapfrog}
& \cite{bae2024singulartrajectory} & \cite{lee2024mart} &  \\
\midrule
1.0s & 0.43/0.66 & 0.26/0.34 & 0.38/0.56 & 0.28/0.37 & 0.35/0.58 & 0.19/0.28 & \underline{0.18}/0.27 & 0.28/0.44 & \underline{0.18}/\underline{0.26} & \textbf{0.17}/\textbf{0.25} \\
2.0s & 0.75/1.24 & 0.49/0.70 & 0.71/1.14 & 0.51/0.72 & 0.68/1.23 & 0.40/0.61 & \underline{0.37}/\underline{0.56} & 0.61/1.00 & \textbf{0.35}/\textbf{0.50} & \textbf{0.35}/\textbf{0.50} \\
3.0s & 1.03/1.51 & 0.73/1.02 & 1.00/1.57 & 0.71/0.98 & 1.01/1.76 & 0.65/0.90 & 0.58/\underline{0.84} & 0.96/1.47 & \underline{0.54}/\textbf{0.71} & \textbf{0.53}/\textbf{0.71} \\
4.0s & 1.13/2.01 & 0.96/1.30 & 1.25/1.47 & 0.96/1.27 & 1.31/1.79 & 0.89/1.13 & 0.81/\underline{1.10} & 1.31/1.98 & \underline{0.73}/\textbf{0.90} & \textbf{0.72}/\textbf{0.90} \\
\bottomrule
\end{tabular}
\label{nba_result}
\endgroup
\end{table*}

\section{EXPERIMENTS}

\subsection{Datasets and Evaluation Metrics}
We evaluate ART on the ETH/UCY \cite{pellegrini2009ETH,Lerner2007UCY} and NBA SportVU \cite{mao2023Leapfrog} benchmarks. The ETH/UCY dataset consists of five subsets, including ETH, HOTEL, UNIV, ZARA1, and ZARA2 captured in diverse social scenarios. Following the standard protocol in \cite{lee2024mart}, we use 3.2 seconds (8 frames) of historical trajectories to predict the subsequent 4.8 seconds (12 frames). The NBA SportVU dataset contains trajectories of 10 players from basketball games. Consistent with \cite{Xu2022GroupNetMH,lee2024mart}, we use 2.0 seconds (10 frames) to predict the next 4.0 seconds (20 frames). We evaluate performance using $\min\mathrm{ADE}_k$ and $\min\mathrm{FDE}_k$, following \cite{gupta2018socialgan,lee2024mart}. Average Displacement Error (ADE) computes the mean Euclidean distance across all predicted time steps, whereas Final Displacement Error (FDE) evaluates the distance at the last time step.

\subsection{Quantitative results}

We evaluate our model against a comprehensive set of SOTA baselines on both ETH/UCY and NBA datasets, including STAR \cite{yu2020spatio}, GroupNet \cite{Xu2022GroupNetMH}, MemoNet \cite{xu2022remembermemo}, MID \cite{gu2022stochastic}, NPSN \cite{bae2022NPSN}, EqMotion \cite{xu2023eqmotion}, ET \cite{bae2023eigentrajectory}, DynGroupNet \cite{xu2024dynamicGroupNet}, LED \cite{mao2023Leapfrog}, SingularTraj \cite{bae2024singulartrajectory}, and MART \cite{lee2024mart}. 

The quantitative results are summarized in Tables \ref{eth_results} and \ref{nba_result}. Our method achieves the best overall results on the ETH/UCY datasets with the lowest average $\min\mathrm{ADE}_{20}/\min\mathrm{FDE}_{20}$ of 0.20/0.32, improving over strong baselines with 0.21 average ADE (e.g., MART/SingularTraj/LED) by 4.8\% while matching the best average FDE. It also attains the best result on ZARA2 (0.12/0.21), demonstrating strong generalization across diverse scenes. On the NBA dataset, our model achieves consistently superior performance across different time scales in terms of both ADE and FDE, highlighting effective long-term interaction modeling and robust prediction accuracy.

\subsection{Ablation Study and Model Analysis}

\begin{figure}[t]\centering
  \includegraphics[width=0.55\linewidth]{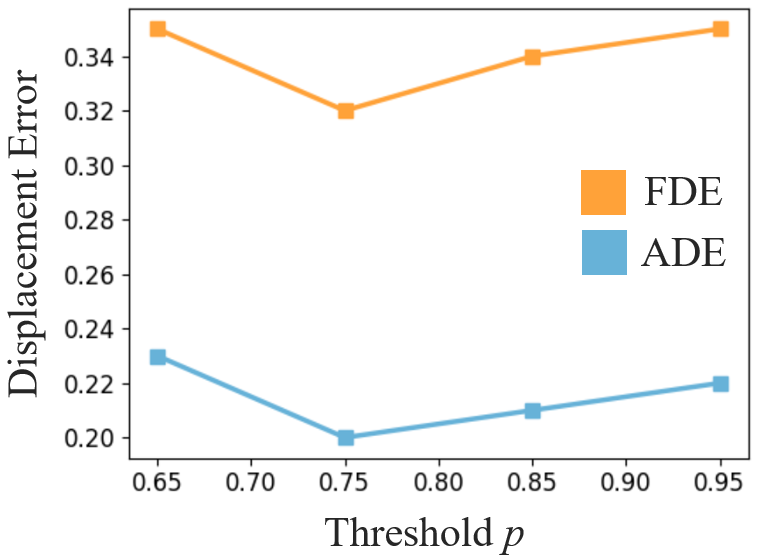}
  \caption{Ablation study of Top-$p$ threshold on the ETH/UCY dataset.}
  \label{fig:threshold-p}
\end{figure}

\subsubsection{Ablation Study on Relation Weighting Strategies}

\begin{table}[t]
\centering
\caption{Ablation study of relation weighting strategies on the ETH/UCY dataset.}
\begingroup
\setlength{\tabcolsep}{4pt}
\small
\begin{tabular}{l|c c}
\toprule
Weighting Strategies & \multicolumn{2}{c}{ETH/UCY Dataset} \\
\cmidrule(lr){2-3}
& $\min\mathrm{ADE}_{20}$ & $\min\mathrm{FDE}_{20}$ \\
\midrule
Cosine Similarity & \underline{0.22} & \underline{0.36} \\
Random Weighting  & 0.23             & 0.37 \\
Uniform Weighting & 0.23             & \underline{0.36} \\
\hline
Ours              & \textbf{0.20}    & \textbf{0.32} \\
\bottomrule
\end{tabular}
\label{ablation_relation_avg_ethucy}
\endgroup
\end{table}

In this section, we evaluate different relation weighting strategies in the proposed Temporal-Aware Relation Graph, including cosine similarity, which assigns static similarity-based weights, random weighting, which ignores temporal structure by randomly sampling weights, and uniform weighting, which applies equal weights to all pairwise relations. As shown in Table~\ref{ablation_relation_avg_ethucy}, our temporally-aware weighting achieves the best performance on the ETH/UCY dataset, demonstrating that explicitly modeling temporal attribution in pairwise relations leads to more effective interaction representations.

\subsubsection{Ablation Study on Top-\textit{p} Threshold}

Figure~\ref{fig:threshold-p} illustrates the impact of different Top-$p$ thresholds on prediction performance. As $p$ decreases from 0.95 to 0.75, both ADE and FDE consistently improve, suggesting that moderate sparsification effectively suppresses weak or noisy interactions while preserving informative relations. When $p=0.95$, the behavior closely resembles the non-sparsified setting, leading to degraded performance due to redundant interaction modeling. Conversely, overly aggressive sparsification ($p=0.65$) also harms performance by discarding useful interactions. Overall, these results demonstrate that a moderate Top-$p$ threshold ($p=0.75$) achieves the best balance between interaction selectivity and information preservation.

\subsubsection{Complexity and Efficiency Analysis}

\begin{figure}[t]
\centering
\includegraphics[width=0.82\columnwidth]{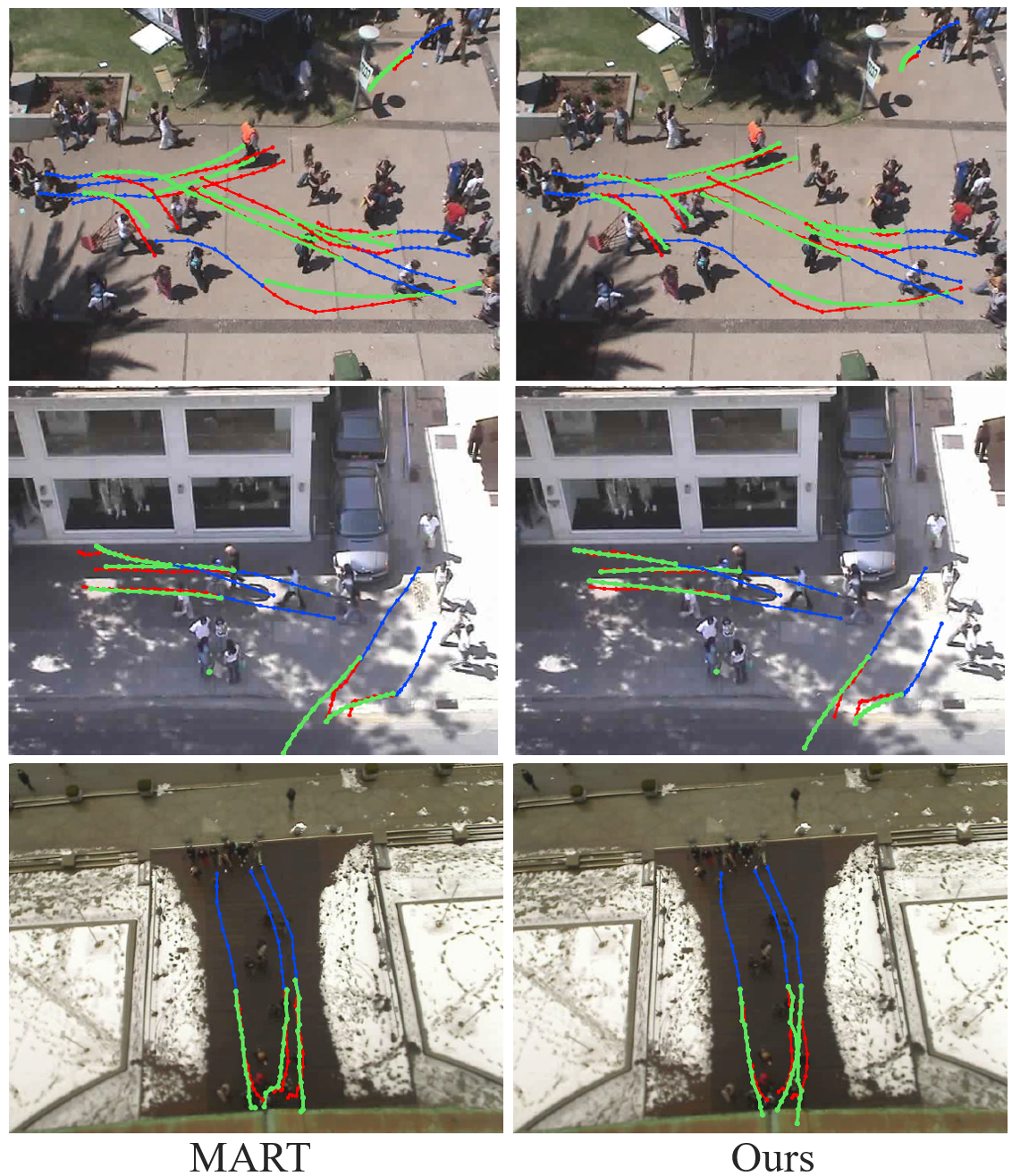} 
\caption{Qualitative comparisons with MART \cite{lee2024mart} on the ETH/UCY dataset. Past trajectories are shown in blue, ground truth in red, and model predictions in green.}
\label{fig:ETHUCY_vis}
\end{figure}

\begin{figure}[t]
\centering
\includegraphics[width=0.85\columnwidth]{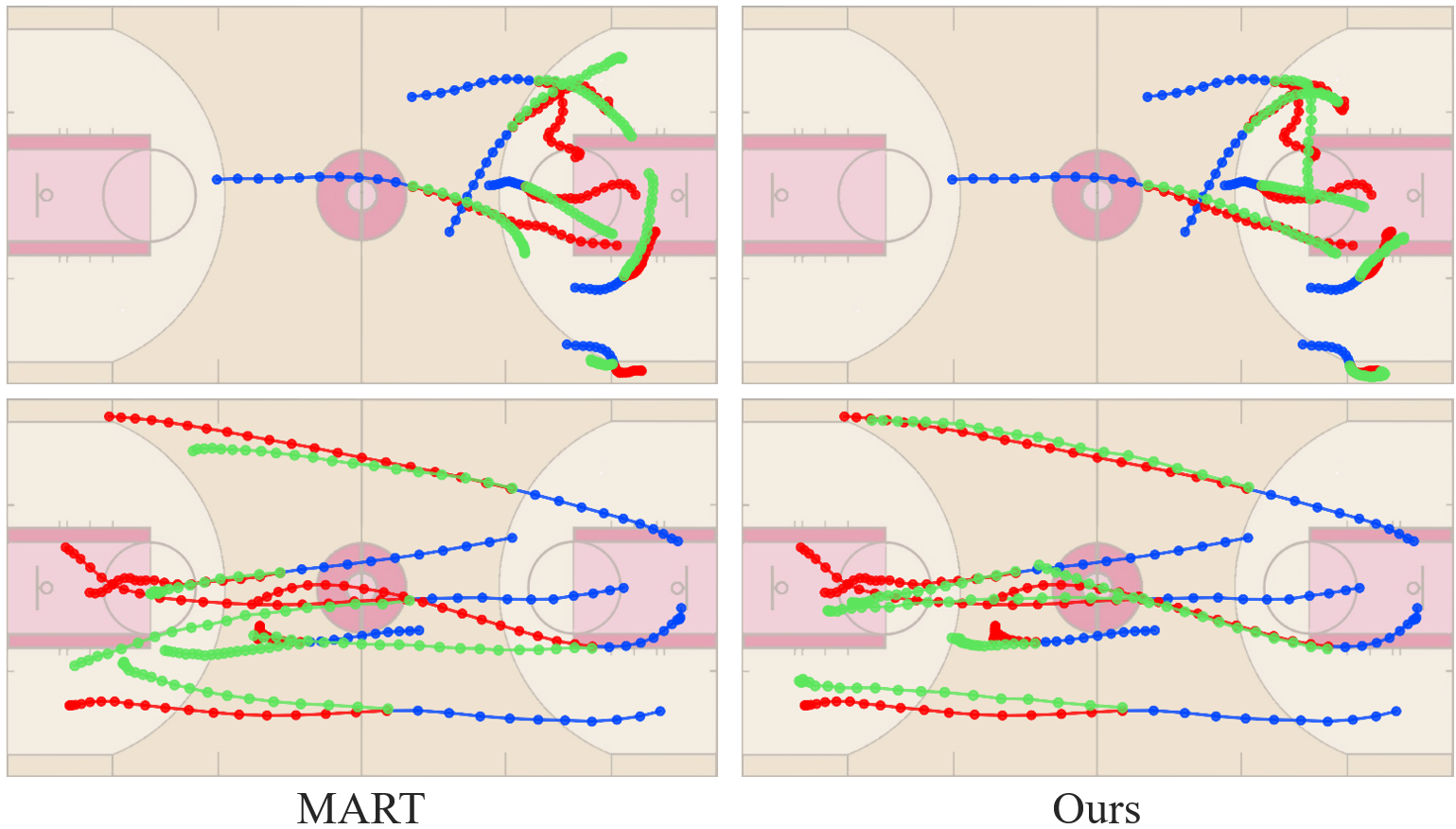} 
\caption{Qualitative comparisons with MART \cite{lee2024mart} on the NBA dataset. Past trajectories are shown in blue, ground truth in red, and predictions in green.}
\label{fig:NBA_vis}
\vspace{-3mm}
\end{figure}

\begin{table}[h]
\centering
\caption{Model complexity comparison. Best and second-best results are marked in \textbf{bold} and \underline{underline}.}
\begingroup
\setlength{\tabcolsep}{3mm}
\small
\begin{tabular}{l|cc}
\toprule
Method       & \#Param. & MACs \\
\midrule
STAR \cite{yu2020spatio}       & \textbf{1.0}M   & 12.0G \\
MemoNet \cite{xu2022remembermemo}    & 10.7M  & 6.0G \\
GroupNet \cite{Xu2022GroupNetMH}    & 2.2M   & 411.5M \\
MID \cite{gu2022stochastic}        & 9.0M   & 40.3G \\
EqMotion \cite{xu2023eqmotion}    & 3.0M   & 147.1M \\
LED \cite{mao2023Leapfrog}        & 10.9M  & 15.0G \\
MART \cite{lee2024mart}       & \underline{1.5}M   & \underline{43.3}M \\
\hline
Ours   & \textbf{1.0}M & \textbf{40.0}M \\
\bottomrule
\end{tabular}
\label{complexity_analysis}
\endgroup
\end{table}

As shown in Table~\ref{complexity_analysis}, our model achieves competitive parameter efficiency with the lowest MACs among all methods, highlighting its efficiency for trajectory prediction. Compared with existing approaches that rely on either larger model sizes or substantially higher computational costs, our method offers a more favorable trade-off between accuracy and efficiency.

\subsection{Qualitative Results}

\subsubsection{Trajectory Prediction Visualizations}

In this section, we present qualitative visualizations to illustrate the superiority of our method. We choose MART \cite{lee2024mart} as the comparison baseline since it is a prior SOTA approach that adopts the same trajectory decoder and relational Transformer architecture, enabling a fair comparison. Figure~\ref{fig:ETHUCY_vis} illustrates the qualitative results on the ETH/UCY dataset across three scenes of increasing interaction complexity, ranging from relatively sparse environments to densely crowded settings. Compared with MART, our predictions more closely follow the ground-truth trajectories and exhibit fewer unrealistic overlaps and collisions, with the advantages becoming more pronounced as scene complexity increases.

Figure~\ref{fig:NBA_vis} presents qualitative comparisons on the NBA dataset. Although long-term trajectory prediction remains challenging for both methods due to highly dynamic and coordinated player movements, our predictions are consistently more aligned with the ground truth than those of MART, demonstrating improved robustness in complex multi-agent sports environments.

\subsubsection{Temporal-Aware Relation Visualization}

\begin{figure}[t]
\centering
\includegraphics[width=0.95\columnwidth]{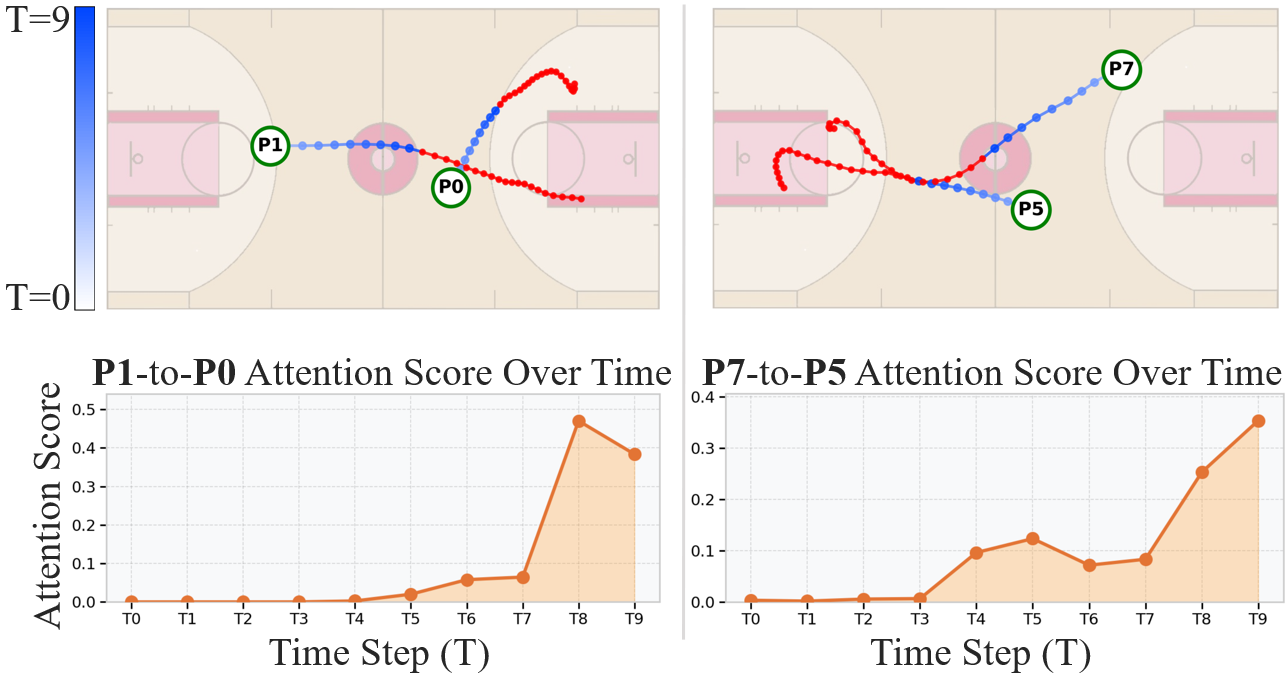} 
\caption{Visualization of temporal-aware relation weights on the NBA dataset. Blue denotes historical trajectories, ground-truth trajectories are in red, and orange indicates the temporal evolution of attention scores.}
\label{fig:attention_vis}
\end{figure}

We qualitatively verify the effectiveness of the proposed TARG in capturing temporally salient interactions. Figure~\ref{fig:attention_vis} provides an intuitive visualization of the proposed TARG on the NBA dataset. By tracking the attention scores between selected player pairs over time, we observe that the model assigns low weights when players are spatially distant and significantly increases the attention as they approach or interact. This behavior indicates that the proposed method can effectively identify and emphasize critical time steps where interactions become salient, thereby preserving temporal attribution rather than uniformly aggregating interactions across time.

\section{CONCLUSION}

We propose ART to model pedestrian interactions through a temporal-aware relation graph and adaptive interaction pruning. By explicitly preserving temporal attribution in pairwise relations and
dynamically sparsifying interaction graphs, ART achieves a favorable balance between expressive interaction modeling and computational efficiency. Our method achieves superior performance on ETH/UCY and NBA, with strong generalization in multi-agent scenarios. Future work will explore more complex and heterogeneous environments using probabilistic models like diffusion models \cite{chang2026design} in real-world robotic systems for interactive decision-making. Other informative features such as whole-body motions \cite{chang2025large} can also be considered in the future.

\bibliographystyle{plain}
\bibliography{reference}

@inproceedings{bhaskara2024trajectory1,
  title={Trajectory prediction for robot navigation using flow-guided markov neural operator},
  author={Bhaskara, Rashmi and Viswanath, Hrishikesh and Bera, Aniket},
  booktitle={ICRA},
  pages={15209--15216},
  year={2024},
  organization={IEEE}
}

@article{akabane2021pedestrian2,
  title={Pedestrian trajectory prediction based on transfer learning for human-following mobile robots},
  author={Akabane, Rina and Kato, Yuka},
  journal={IEEE Access},
  volume={9},
  pages={126172--126185},
  year={2021},
  publisher={IEEE}
}

@ARTICLE{huang2022pedestrian3,
  author={Huang, Zhe and Li, Ruohua and Shin, Kazuki and Driggs-Campbell, Katherine},
  journal={IEEE RAL}, 
  title={Learning Sparse Interaction Graphs of Partially Detected Pedestrians for Trajectory Prediction}, 
  year={2022},
  volume={7},
  number={2},
  pages={1198-1205},}

@inproceedings{bai2015pomdpintro,
  title={Intention-aware online POMDP planning for autonomous driving in a crowd},
  author={Bai, Haoyu and Cai, Shaojun and Ye, Nan and Hsu, David and Lee, Wee Sun},
  booktitle={ICRA},
  pages={454--460},
  year={2015},
}

@article{luo2018porca,
  title={Porca: Modeling and planning for autonomous driving among many pedestrians},
  author={Luo, Yuanfu and Cai, Panpan and Bera, Aniket and Hsu, David and Lee, Wee Sun and Manocha, Dinesh},
  journal={IEEE RAL},
  volume={3},
  number={4},
  pages={3418--3425},
  year={2018},
}

@article{li2025unifiedTraj,
  title={Unified Spatial-Temporal Edge-Enhanced Graph Networks for Pedestrian Trajectory Prediction},
  author={Li, Ruochen and Qiao, Tanqiu and Katsigiannis, Stamos and Zhu, Zhanxing and Shum, Hubert PH},
  journal={TCSVT},
  year={2025},
}

@article{chen2018robotnavigation1,
  title={Robot navigation based on human trajectory prediction and multiple travel modes},
  author={Chen, Zhixian and Song, Chao and Yang, Yuanyuan and Zhao, Baoliang and Hu, Ying and Liu, Shoubin and Zhang, Jianwei},
  journal={Applied Sciences},
  volume={8},
  number={11},
  pages={2205},
  year={2018},
  publisher={MDPI}
}

@article{yang2024ia,
  title={IA-LSTM: Interaction-aware LSTM for pedestrian trajectory prediction},
  author={Yang, Jing and Chen, Yuehai and Du, Shaoyi and Chen, Badong and Principe, Jose C},
  journal={IEEE transactions on cybernetics},
  volume={54},
  number={7},
  pages={3904--3917},
  year={2024},
  publisher={IEEE}
}

@inproceedings{Alexandre2016lstm,
  title={Social lstm: Human trajectory prediction in crowded spaces},
  author={Alahi, Alexandre and Goel, Kratarth and Ramanathan, Vignesh and Robicquet, Alexandre and Fei-Fei, Li and Savarese, Silvio},
  booktitle={CVPR},
  pages={961--971},
  year={2016}
}

@inproceedings{gupta2018socialgan,
  title={Social gan: Socially acceptable trajectories with generative adversarial networks},
  author={Gupta, Agrim and Johnson, Justin and Fei-Fei, Li and Savarese, Silvio and Alahi, Alexandre},
  booktitle={CVPR},
  pages={2255--2264},
  year={2018}
}

@inproceedings{xue2018ss-lstm,
  title={SS-LSTM: A hierarchical LSTM model for pedestrian trajectory prediction},
  author={Xue, Hao and Huynh, Du Q and Reynolds, Mark},
  booktitle={WACV},
  pages={1186--1194},
  year={2018},
}

@inproceedings{guo2022end2end,
  title={End-to-end trajectory distribution prediction based on occupancy grid maps},
  author={Guo, Ke and Liu, Wenxi and Pan, Jia},
  booktitle={CVPR},
  pages={2242--2251},
  year={2022}
}

@inproceedings{ruochen2022multiclassSGCN,
  title={Multiclass-SGCN: Sparse Graph-Based Trajectory Prediction with Agent Class Embedding},
  author={Li, Ruochen and Katsigiannis, Stamos and Shum, Hubert PH},
  booktitle={ICIP},
  pages={2346--2350},
  year={2022},
  organization={IEEE}
}

@article{li2025bpsgcn,
  title={BP-SGCN: Behavioral Pseudo-Label Informed Sparse Graph Convolution Network for Pedestrian and Heterogeneous Trajectory Prediction},
  author={Li, Ruochen and Katsigiannis, Stamos and Kim, Tae-Kyun and Shum, Hubert PH},
  journal={TNNLS},
  year={2025},
}

@article{Xu2022GroupNetMH,
  title={GroupNet: Multiscale Hypergraph Neural Networks for Trajectory Prediction with Relational Reasoning},
  author={Chenxin Xu and Maosen Li and Zhenyang Ni and Ya Zhang and Siheng Chen},
  journal={CVPR},
  year={2022},
  pages={6488-6497},
}

@inproceedings{xu2023eqmotion,
      title={{EqMotion}: Equivariant Multi-Agent Motion Prediction With Invariant Interaction Reasoning}, 
      author={Chenxin Xu and Robby T. Tan and Yuhong Tan and Siheng Chen and Yu Guang Wang and Xinchao Wang and Yanfeng Wang},
      year={2023},
      booktitle = {CVPR}
}

@inproceedings{lee2024mart,
  title={Mart: Multiscale relational transformer networks for multi-agent trajectory prediction},
  author={Lee, Seongju and Lee, Junseok and Yu, Yeonguk and Kim, Taeri and Lee, Kyoobin},
  booktitle={ECCV},
  pages={89--107},
  year={2024},
}

@inproceedings{shi2023tutr,
  title={Trajectory unified transformer for pedestrian trajectory prediction},
  author={Shi, Liushuai and Wang, Le and Zhou, Sanping and Hua, Gang},
  booktitle={ICCV},
  pages={9675--9684},
  year={2023}
}

@inproceedings{yu2020spatio,
  title={Spatio-temporal graph transformer networks for pedestrian trajectory prediction},
  author={Yu, Cunjun and Ma, Xiao and Ren, Jiawei and Zhao, Haiyu and Yi, Shuai},
  booktitle={ECCV},
  pages={507--523},
  year={2020},
  organization={Springer}
}

@inproceedings{yuan2021agentformer,
  title={Agentformer: Agent-aware transformers for socio-temporal multi-agent forecasting},
  author={Yuan, Ye and Weng, Xinshuo and Ou, Yanglan and Kitani, Kris M},
  booktitle={ICCV},
  pages={9813--9823},
  year={2021}
}

@article{vaswani2017transformer,
  title={Attention is all you need},
  author={Vaswani, Ashish and Shazeer, Noam and Parmar, Niki and Uszkoreit, Jakob and Jones, Llion and Gomez, Aidan N and Kaiser, {\L}ukasz and Polosukhin, Illia},
  journal={Neurips},
  volume={30},
  year={2017}
}

@inproceedings{xu2022remembermemo,
  title={Remember intentions: Retrospective-memory-based trajectory prediction},
  author={Xu, Chenxin and Mao, Weibo and Zhang, Wenjun and Chen, Siheng},
  booktitle={CVPR},
  pages={6488--6497},
  year={2022}
}

@inproceedings{bae2022NPSN,
  title={Non-probability sampling network for stochastic human trajectory prediction},
  author={Bae, Inhwan and Park, Jin-Hwi and Jeon, Hae-Gon},
  booktitle={CVPR},
  pages={6477--6487},
  year={2022}
}

@article{xu2024dynamicGroupNet,
  title={Dynamic-group-aware networks for multi-agent trajectory prediction with relational reasoning},
  author={Xu, Chenxin and Wei, Yuxi and Tang, Bohan and Yin, Sheng and Zhang, Ya and Chen, Siheng and Wang, Yanfeng},
  journal={Neural Networks},
  volume={170},
  pages={564--577},
  year={2024},
}

@inproceedings{bae2023eigentrajectory,
  title={Eigentrajectory: Low-rank descriptors for multi-modal trajectory forecasting},
  author={Bae, Inhwan and Oh, Jean and Jeon, Hae-Gon},
  booktitle={ICCV},
  pages={10017--10029},
  year={2023}
}

@inproceedings{gu2022stochastic,
  title={Stochastic trajectory prediction via motion indeterminacy diffusion},
  author={Gu, Tianpei and Chen, Guangyi and Li, Junlong and Lin, Chunze and Rao, Yongming and Zhou, Jie and Lu, Jiwen},
  booktitle={CVPR},
  pages={17113--17122},
  year={2022}
}

@InProceedings{mao2023Leapfrog,
      title={Leapfrog Diffusion Model for Stochastic Trajectory Prediction}, 
      author={Weibo Mao and Chenxin Xu and Qi Zhu and Siheng Chen and Yanfeng Wang},
      year={2023},
      booktitle = {CVPR}
}

@inproceedings{bae2024singulartrajectory,
  title={Singulartrajectory: Universal trajectory predictor using diffusion model},
  author={Bae, Inhwan and Park, Young-Jae and Jeon, Hae-Gon},
  booktitle={CVPR},
  pages={17890--17901},
  year={2024}
}

@article{ha2022learningrain,
  title={Learning heterogeneous interaction strengths by trajectory prediction with graph neural network},
  author={Ha, Seungwoong and Jeong, Hawoong},
  journal={arXiv},
  year={2022}
}

@article{li2025vite,
  title={ViTE: Virtual Graph Trajectory Expert Router for Pedestrian Trajectory Prediction},
  author={Li, Ruochen and Zhu, Zhanxing and Qiao, Tanqiu and Shum, Hubert PH},
  journal={arXiv},
  year={2025}
}

@inproceedings{
lin2025twilight,
title={Twilight: Adaptive Attention Sparsity with Hierarchical Top-\$p\$  Pruning},
author={Chaofan Lin and Jiaming Tang and Shuo Yang and Hanshuo Wang and Tian Tang and Boyu Tian and Ion Stoica and Song Han and Mingyu Gao},
booktitle={Neurips},
year={2025},
}

@inproceedings{pellegrini2009ETH,
  title={You'll never walk alone: Modeling social behavior for multi-target tracking},
  author={Pellegrini, Stefano and Ess, Andreas and Schindler, Konrad and Van Gool, Luc},
  booktitle={ICCV},
  pages={261--268},
  year={2009},
}

@inproceedings{Lerner2007UCY,
  title={Crowds by example},
  author={Lerner, Alon and Chrysanthou, Yiorgos and Lischinski, Dani},
  booktitle={Computer graphics forum},
  volume={26},
  pages={655--664},
  year={2007}}

@inproceedings{
petar2018GAT,
title={Graph attention networks},
author={Veli{\v{c}}kovi{\'c}, Petar and Cucurull, Guillem and Casanova, Arantxa and Romero, Adriana and Lio, Pietro and Bengio, Yoshua},
booktitle={ICLR},
year={2018},
}

@inproceedings{
diao2023relational,
title={Relational Attention: Generalizing Transformers for Graph-Structured Tasks},
author={Cameron Diao and Ricky Loynd},
booktitle={ICLR},
year={2023},
}

@inproceedings{qiao2024category,
  title={From category to scenery: An end-to-end framework for multi-person human-object interaction recognition in videos},
  author={Qiao, Tanqiu and Li, Ruochen and Li, Frederick WB and Shum, Hubert PH},
  booktitle={ICPR},
  pages={262--277},
  year={2024},
}

@article{qiao2025geometric,
  title={Geometric visual fusion graph neural networks for multi-person human-object interaction recognition in videos},
  author={Qiao, Tanqiu and Li, Ruochen and Li, Frederick WB and Kubotani, Yoshiki and Morishima, Shigeo and Shum, Hubert PH},
  journal={arXiv},
  year={2025}
}

@inproceedings{li2026vite,
  title={ViTE: Virtual Graph Trajectory Expert Router for Pedestrian Trajectory Prediction},
  author={Li, Ruochen and Zhu, Zhanxing and Qiao, Tanqiu and Shum, Hubert PH},
  booktitle={AAAI},
  volume={40},
  number={21},
  pages={17535--17543},
  year={2026}
}

@article{chang2026design,
  title={On the design fundamentals of diffusion models: A survey},
  author={Chang, Ziyi and Koulieris, George A and Chang, Hyung Jin and Shum, Hubert PH},
  journal={Pattern Recognition},
  volume={169},
  pages={111934},
  year={2026},
  publisher={Elsevier}
}

@inproceedings{chang2025large,
  title={Large-scale multi-character interaction synthesis},
  author={Chang, Ziyi and Wang, He and Koulieris, George and Shum, Hubert PH},
  booktitle={Proceedings of the Special Interest Group on Computer Graphics and Interactive Techniques Conference Conference Papers},
  pages={1--10},
  year={2025}
}

\end{document}